\pdfoutput=1

\documentclass[11pt]{article}

\usepackage[]{acl}
\usepackage{comment}

\usepackage{times}
\usepackage{latexsym}
\usepackage{float}

\usepackage{graphicx}
\usepackage{multirow}
\usepackage{multicol}
\usepackage{tipa}
\usepackage{amsmath}

\usepackage{booktabs}
\usepackage{dcolumn}
\newcolumntype{d}{D{.}{.}{-1}}

\usepackage[T1]{fontenc}

\usepackage[utf8]{inputenc}

\usepackage{microtype}

\newcommand{\newpar}[1]{\noindent\textbf{{#1}~}}

\title{LAIT: Efficient Multi-Segment Encoding in Transformers with\\ Layer-Adjustable Interaction}

\author{Jeremiah Milbauer$^{1,2}$\thanks{\quad Work done as an intern at Google Research.}, Annie Louis$^{1}$, Mohammad Javad Hosseini$^{1}$, \\ \textbf{Alex Fabrikant$^{1}$, Donald Metzler$^{1}$, Tal Schuster$^{1}$}
\\
\\
$^1$Google Research \qquad $^2$Carnegie Mellon University \\
{\tt jmilbaue@andrew.cmu.edu, talschuster@google.com}}

\begin{document}
\maketitle
\begin{abstract}

Transformer encoders contextualize token representations by attending to all other tokens at each layer, leading to quadratic increase in compute effort with the input length. In practice, however, the input text of many NLP tasks can be seen as a sequence of related segments (e.g., the sequence of sentences within a passage, or the hypothesis and premise in NLI). While attending across these segments is highly beneficial for many tasks, we hypothesize that this interaction can be delayed until later encoding stages.
To this end, we introduce \textbf{L}ayer-\textbf{A}djustable \textbf{I}nteractions in \textbf{T}ransformers (LAIT). Within LAIT, segmented inputs are first encoded independently, and then jointly. This partial two-tower architecture bridges the gap between a Dual Encoder's ability to pre-compute representations for segments and a fully self-attentive Transformer's capacity to model cross-segment attention. The LAIT framework effectively leverages existing pretrained Transformers and converts them into the hybrid of the two aforementioned architectures, allowing for easy and intuitive control over the performance-efficiency tradeoff. Experimenting on a wide range of NLP tasks, we find LAIT able to reduce 30-50\% of the attention FLOPs on many tasks, while preserving high accuracy; in some practical settings, LAIT could reduce actual latency by orders of magnitude.

\end{abstract}

\section{Introduction}
\label{sec:intro}

\begin{figure}[t]
    \centering
    \includegraphics[width=\linewidth]{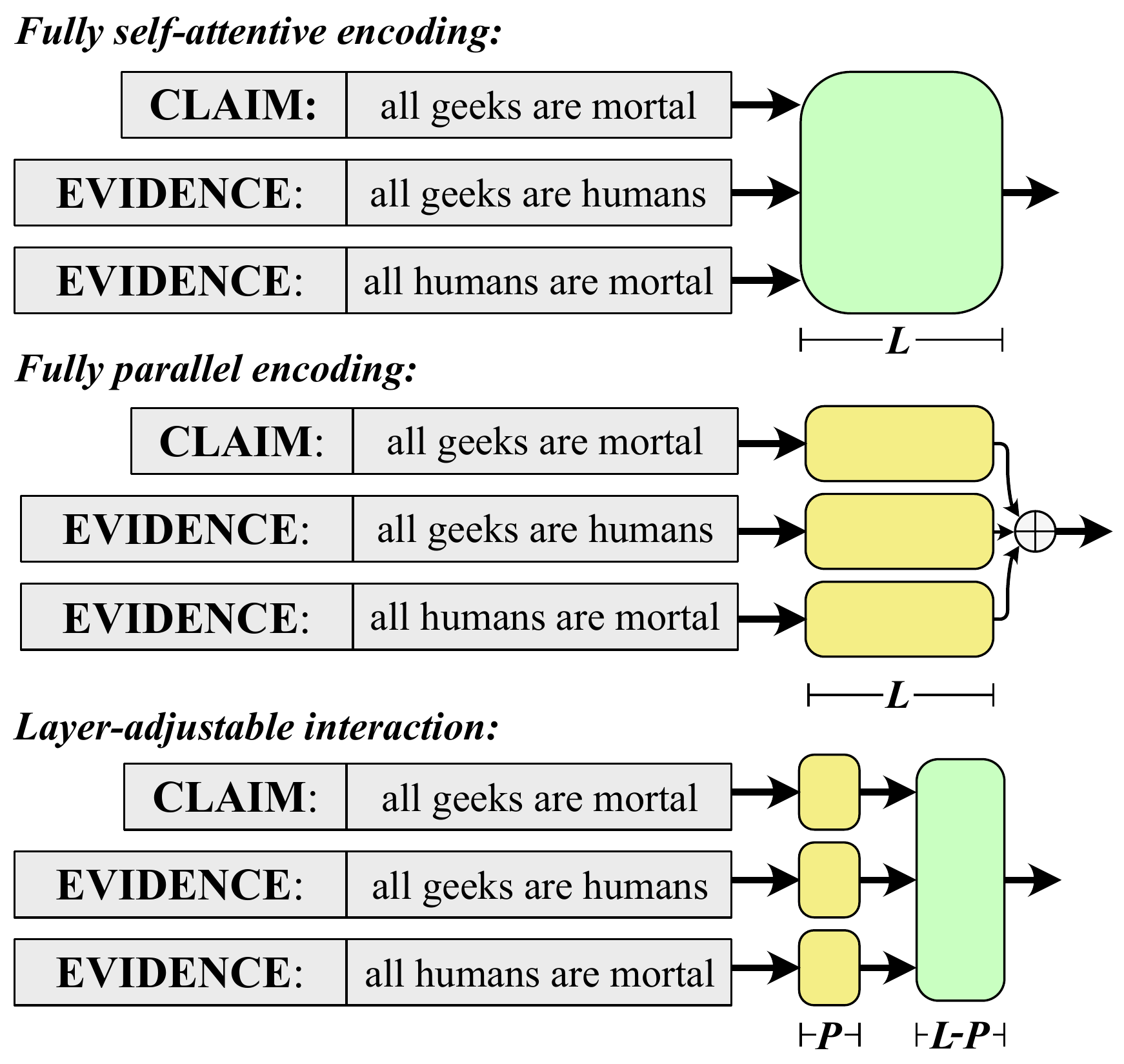}
    \caption{A comparison of three approaches to multi-segment modeling for an arbitrary claim verification task. a) Fully-self attentive architecture, with each token attending to each other token over $L$ layers. b) Generalized dual encoder, with each segment encoded separately by an $L$-layer Transformer and representations concatenated. c) Layer-adjustable interactions (\textbf{ours}), with N layers of independent segment encoding and $L-P$ layers of fully self-attentive segment encoding.}
    \label{fig:comparison_of_architectures}
\end{figure}

Although the meaning of a sentence may depend on the context in which it appears, sentences still have meaning \textit{per se}. However, in tasks involving reasoning across multiple sentences or text segments --- like natural language inference (NLI), fact verification, question answering (QA), semantic similarity (STS), etc. --- the common setting is to concatenate and \textit{jointly} process all tokenized segments as input to a neural model, most often some form of bidirectional Transformer-based architecture~\citep{transformer}. In this setting, the self-attention blocks of the Transformer layers contextualize the per-token representations against all other input tokens, including those of different input segments. The potential for independent sentence-level semantics is largely ignored.

While this practice has shown to achieve high accuracy, it is computationally expensive due to the quadratic increase in cost with the input length. And in practical settings, such as large-scale citation retrieval \cite{petroni2022improving} or document-level NLI \cite{koreeda2021contractnli}, where a given segment may occur multiple times, the full Cartesian product of the sets of text segments must be processed, e.g., \newcite{schuster2022stretching} processes all sentence pairs from two Wikipedia articles around one subject but in two different languages to identify potential discrepancies. This leads to yet another quadratic increase in cost. Our goal is to reduce both of these computational burdens, rendering transformer architectures more efficient for large-scale multi-segment reasoning.

In this paper, we present LAIT (/le\textsci{}t/), a late interaction Transformer model with easy to implement \textbf{L}ayer-\textbf{A}djustable \textbf{I}nteractions. LAIT includes encoder layers that process each segment locally and independent of the other segments, followed by traditional Transformer layers, in a simple but effective way. Unlike the late interaction components of other models, such as ColBERT \cite{khattab2020colbert}, which are specifically geared toward measuring a similarity score between two text segments, LAIT generally supports any sequence-to-sequence task and any number of input segments.

LAIT enables several desirable properties for an efficient encoder: it (1) is easy to train on top of existing pretrained language models; (2) readily supports any seq-2-seq task, and any segmentation of the input; (3) improves the encoding efficiency by skipping a large number of attention computations; (4) disentangles independent segment representations from joint processing to allow caching of intermediate segment representations for repeated computations; and (5) provides an easy-to-tune hyperparameter for controlling the efficiency-performance tradeoff.

\section{Background: Full Self-attention vs.\ Dual Encoders}
\label{sec:motivation}

A key strength of a fully self-attentive (FSA) architecture, such as BERT or T5 \citep{devlin-etal-2019-bert, raffel2020exploring} is the ability of each token in the input to interact with each other token in the input throughout all layers of the model. Although expensive, this type of architecture has shown impressive performance across a wide variety of NLP tasks such as those in the GLUE and SuperGLUE benchmarks \citep{wang2019glue, wang2019superglue}.

A common alternative to FSA is the dual encoder (DE) framework \cite{gillick2018end}. With DE, two text segments are embedded independently, either by separate networks or by two networks that share parameters. A DE typically involves two encoders, $\textnormal{Enc}_{q}(\cdot)$ and $\textnormal{Enc}_{d}(\cdot)$, and a comparison function $\textnormal{Comp}(\cdot)$, and for a given pair of input segments $q, d$:
$
   \textnormal{score} = \textnormal{Comp}(\textnormal{Enc}_q(q), \textnormal{Enc}_d(d)).
  $
In practice, the two encoders can share parameters.

DE is typically trained with a contrastive loss over a set of positive $q,d$ pairs, with the goal of having the score of positive pairs greater than that of negatives. Therefore, DE is most suited for similarity tasks such as information retrieval.

A specific advantage of the DE architecture for retrieval tasks is its ability to independently encode the two input segments. In practice, this allows encoding and storing many documents' representations in parallel in advance. Then, only new queries need to be encoded into a vector that can be used for retrieving the top similar documents from the pre-encoded corpus using efficient methods such as maximum inner product search (MIPS).

The method above, however, only supports similarity tasks or binary classification tasks over input pairs. To expand this setting to multi-class tasks, prior approaches like \citet{casanueva2020, ni-etal-2022-sentence} add a classification head with optional non-linear layers on top of the two encoded representations. Since the classifier requires a fixed-size input, the segment representations are aggregated (e.g., by taking the average over tokens, or by selecting a predefined special token). While conceptually enabling any classification task, the performance of such models is usually far behind the state-of-the-art (see Section~\ref{sec:results}).

\section{Layer-Adjustable Interactions}
\label{sec:model}

\begin{figure*}[t]
    \centering
    \includegraphics[width=0.85\textwidth]{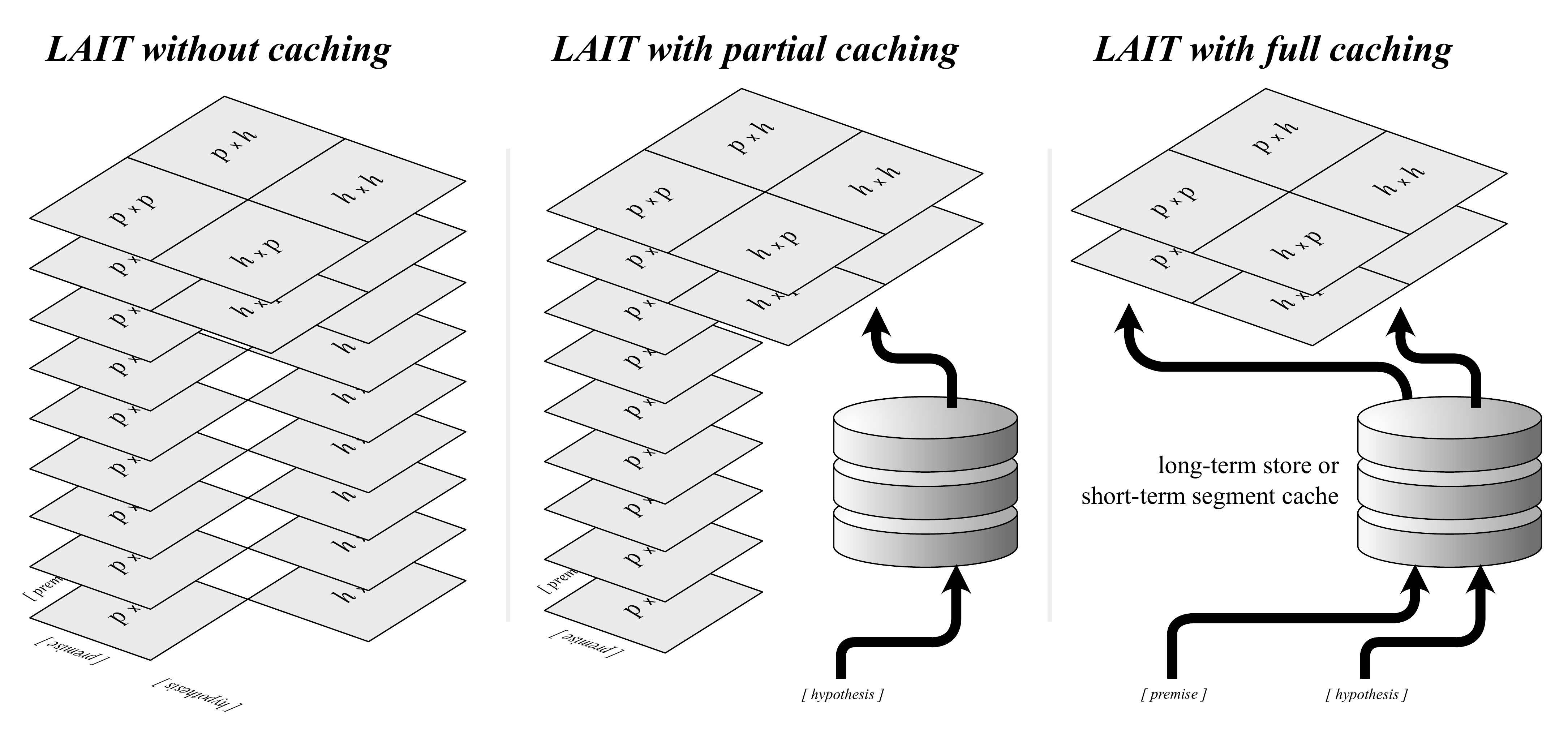}
    \caption{Depiction of a 9-layer LAIT architecture for a 2 segment task (such as NLI) with 7 parallel layers and 2 fully self-attentive layers. Without caching, LAIT reduces computation by eliminating cross-attention for 7 layers. With partial or full caching of segments, LAIT achieves further reductions by re-using independently encoded segment representations.}
    \label{fig:architecture}
    \vspace{-5pt}
\end{figure*}

We argue
that both FSA and DE Transformer models can be seen as special cases of a general architecture with adjustable layer depths for both segment-independence and segment-interaction, which we will call a ``\textbf{L}ayer-\textbf{A}djustable \textbf{I}nteraction \textbf{T}ransformer" (LAIT).

For a Transformer with $L$ layers and an input with $N$ segments, LAIT is a set of $N$ independent stacks of $P$ layers each, followed by $L-P$ fully self-attentive encoder layers. Any function can be used after the encoder. Thus a typical fully self-attentive Encoder-Decoder Transformer is a LAIT where $P=0$, and a shared-parameter dual encoder is a LAIT where $P=L$ and $N=2$. In the fully self-attentive Transformer, each token in each segment is interacting with each token in each other segment throughout the entire depth of the encoder; in a Dual Encoder, each segment is treated independently throughout the encoder.

The LAIT framework allows us to make the core questions of this work precise: (1) to what extent are interactions across multiple input text segments necessary? And (2) If they are not always necessary, how can we take advantage of this fact to perform multi-segment modeling efficiently at scale?

Specifically, given an input $X$ with $m$ tokens that is split into $n$ segments $s_i \ldots s_n$ of possibly different lengths, the LAIT encoder is defined as:
\begin{align*}
\begin{split}
 &\textnormal{LAIT}(s_1, s_2, \ldots, s_n) =\\ &\textnormal{Enc}_{L-P}(
[\textnormal{Enc}_P(s_1) ;
\textnormal{Enc}_P(s_2) ;
\ldots ;
\textnormal{Enc}_P(s_n) ]
),
\end{split}
\end{align*}
where $[x;y]$ denotes concatenating vectors $x$ and $y$, and $\textnormal{Enc}_K(\cdot)$ denotes a Transformer encoder with $K$ layers.

The rule for splitting the input into segments $R(x_1,\ldots,x_m) \rightarrow s_1, \ldots, s_n$ is predefined for each task, based either on prior knowledge of the input structure, or on a simple segmentation function. For example, in NLI we can simply use the hypothesis and premise as two segments. In passage-level QA, we can use the question as one segment and the passage as another. However, splitting the passage into multiple shorter segments could help further reduce compute. For instance, we can split the passage by sentences to $k$ segments, leading to a total of $k+1$ segments.

For $P \in [0, L]$, LAIT interpolates between an $N$-Encoder model and a fully-self attentive Transformer. Because interaction between segments is delayed, representations computed at layer $P$ of the model can be stored or cached for later reuse as they are independently generated. Figure \ref{fig:architecture} demonstrates the basic LAIT architecture, as well as possibilities for partial caching (for instance, multiple unique questions about the same passage), or full caching (for instance, NLI-based cross-document reasoning~\citep{schuster2022stretching}).

Similar to general text-to-text models, the outputs of the LAIT encoder, consisting of $m$ contextualized representations for $m$ tokens, are passed to the Transformer-decoder for generating the output sequence. Similarly, the decoder may be replaced with a classification head, or any other module.

\subsection{Attention Complexity}

By first processing text independently, and then processing the intermediate representations jointly, LAIT reduces the attention complexity within a Transformer in accordance with both the degree of independence (i.e., $P$) and the balance of length across segment inputs. We can calculate the number of attention operations, $\mathcal{O}$, for a given input to LAIT with the formula:

\begin{align}
    \mathcal{O} &= \mathcal{O}_{\textnormal{PAR}} + \mathcal{O}_{\textnormal{FSA}} \label{eq:op_formula} \\
    \mathcal{O}_{\textnormal{PAR}} &= P \cdot \sum_{i=1}^{n} |s_i|^2 \\
    \mathcal{O}_{\textnormal{FSA}} &= (L-P) \cdot \Bigl[ \sum_{i=1}^{n} |s_i| \Bigr]^2
\end{align}
where $|s_i|$ denotes the length of segment $i$ out of $n$ total segments for a given input.

Ultimately, the number of FLOPs to process a single example will depend on the lengths of the input segments, the Transformer architecture used, and the degree of independence $P$. We discuss these practical details in Section \ref{sec:tasks}, and Table \ref{tab:95_flops_table}.

\begin{figure}[t]
    \centering
    \includegraphics[width=\linewidth]{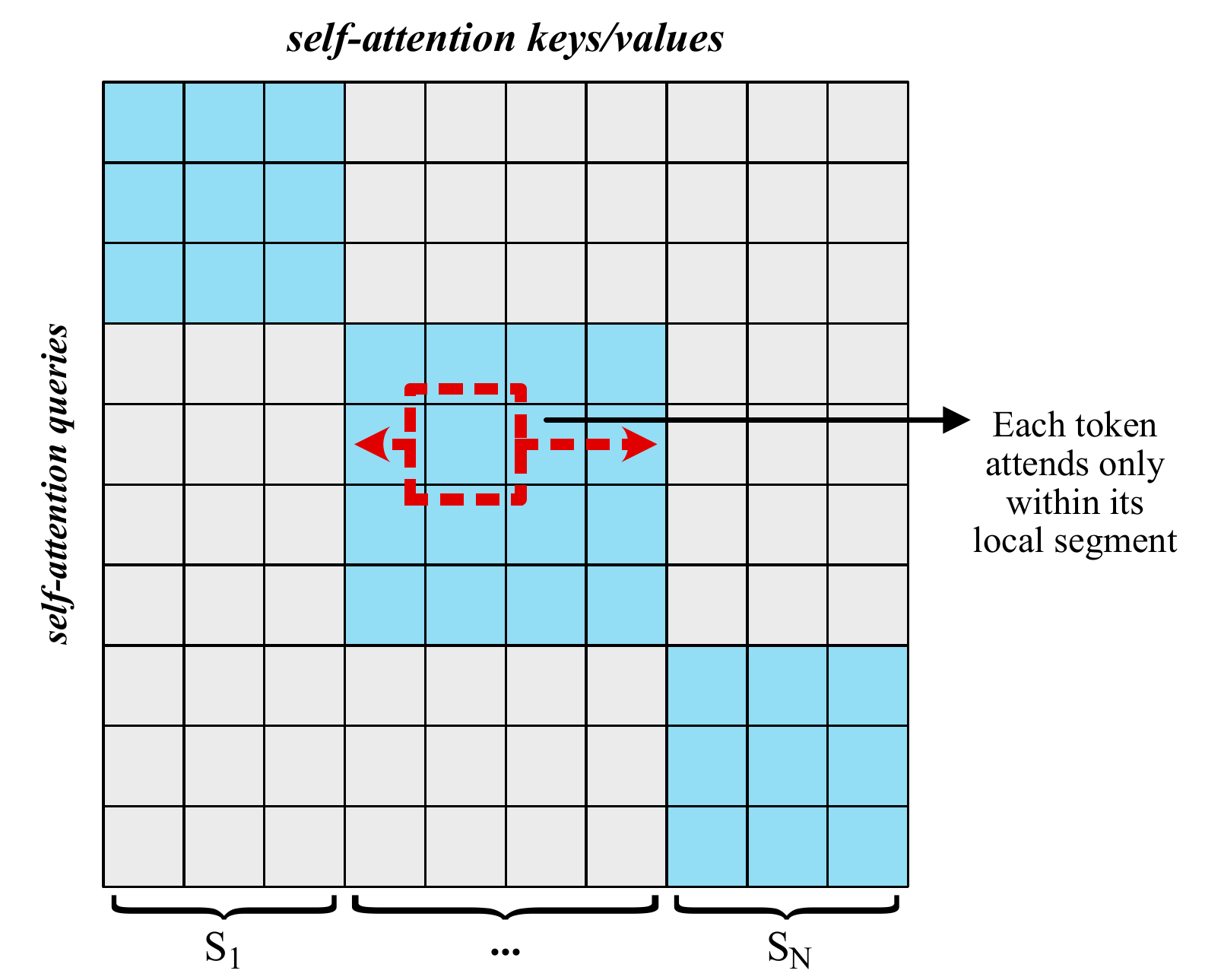}
    \caption{In the parallel layers of LAIT, segments are concatenated but a block-diagonal attention mask maintains independent encoding of each segment. Figure design adapted from \citet{guo2021longt5}.}
    \label{fig:block-diagonal}
\end{figure}

\subsection{Training LAIT}\label{sec:training_lait}
Thanks to LAIT not adding any new parameters to the Transfomer architecture, we can easily convert an existing Transformer to the LAIT framework and train it end-to-end with any objective. In this work, we focus on the T5~\cite{raffel2020exploring} model since it is a general text-to-text Transfomer, and apply LAIT to the encoder stack. In our experiments here, since we focus on classification tasks, we only keep a single decoding layer.

Given an input with $n$ text segments, LAIT first encodes and concatenates the segments. During encoding, a block-diagonal attention mask restricts attention between different text segments for the early layers of the model (denoted ``parallel layers"), and allows cross-segment attention for the later layers of the model (``joint layers"). Figure \ref{fig:block-diagonal} illustrates the block-diagonal attention mask used for parallel layers.

This approach allows for parameter sharing while independently encoding the segments, as well as flexibility for tasks with different numbers of input segments without needing to initialize additional models.

\section{Experimental Setting}
\label{sec:experiments}

Below, we describe our evaluation setting, tasks, used metrics, and baselines.
\subsection{Implementation details}

We implement LAIT on top of the T5 model~\cite{raffel2020exploring} using Google's T5x library~\citep{roberts2022t5x}. In all experiments, we use \texttt{T5-base} which has a total of 12 encoder layers and 220M parameters. To reduce compute effort, we use only a single decoder layer for LAIT (See Appendix~\ref{sec:appendix_larger_models} for larger models). We load the parameters from the public pretrained checkpoint, and finetune on the target task for up to 100K steps with different LAIT configurations (value of $P$). We train LAIT on 16 TPUv3 chips, taking about 4 hours per run. We run a small grid search over learning rate and batch size configurations, and pick the top performing checkpoint based on validation performance.

\subsection{Tasks and metrics}
\label{sec:tasks}

We experiment using LAIT on a diverse set of common tasks and datasets. For each task, we must determine which fields of the dataset to use as input segments for LAIT. We evaluate each task using its typical quality metric. In addition, to measure the efficiency gains of different LAIT configurations, we compute the average self-attention FLOPs. We use Equation (\ref{eq:op_formula}) and the precise configuration of the \texttt{T5-base} model we implement LAIT within, which has $768$-dimensional embeddings and $12$ $64$-dimensional attention heads. %

The evaluated tasks are described below. Many of these tasks are from the popular GLUE~\cite{wang2019glue} and SuperGLUE~\cite{wang2019superglue} benchmarks, and all are in English.  Number of used segments and average lengths per task are summarized in Table~\ref{tab:tasks_stats}. Pre-processing and concatenation strategy are described in Appendix~\ref{sec:concat_strategy}.

\newpar{MNLI} \citep{williams2018mnli}: A dataset for natural language inference across diverse categories. We use the hypothesis and premise as separate segments, and predict one of three labels: ``entailment", ``contradiction", and ``neutral". We report accuracy on the ``matched'' eval set.

\noindent\textbf{RTE}: The Recognizing Textual Entailment dataset combines the data from a series of annual textual entailment challenges \cite{dagan2005pascal,bar2006second,giampiccolo2007third,bentivogli2009fifth}. We use the hypothesis and the premise as separate segments and predict ``entailment'' vs.\ ``non-entailment'', and measure accuracy.

\noindent\textbf{QQP} \citep{WinNT}: Quora Question Pairs dataset is a collection of question pairs from Quora, where the task is to determine whether a pair of questions have the same meaning. For LAIT, we treat each question as a segment, and predict ``duplicate'' or ``not\_duplicate'', and measure accuracy.

\newpar{STSB} \citep{cer2017stsb}: Semantic textual similarity benchmark, a task for estimating the similarity of a pair of sentences. We use each sentence as a separate segment, and predict a score in $[0, 5]$, represented as a string rounded to 2 decimal places. We measure Spearman correlation.

\begin{table}[t]
    \centering
    \resizebox{\linewidth}{!}{%
    \begin{tabular}{l|c|l}
    \toprule
    Task & $n$ & Avg.\ segment lengths\\
    \midrule
         MNLI & 2 & hyp.: 16.14, prem.: 30.79  \\
         RTE & 2 & hyp.: 9.40, prem.: 43.39\\
         QQP & 2 & q.1: 11.94, q.2: 12.17\\
         STSB & 2 & sent.1: 19.71, sent.2: 19.75 \\
         AE & 3 & cand.: 6.80, ref.: 6.12, q.: 12.10  \\
         BoolQ & 2 & pass.: 135.82, q.: 14.54 \\
         BoolQ-Split & 6 & pass.1-5: 29.57, q.: 14.54 \\
         WiC & 2 & w.+sent.1: 14.69, w.+sent.2: 14.88\\
         FEVER & 2 & claim: 15.90, evid.: 46.20 \\
         VitaminC & 2 & claim: 21.43, evid.: 43.78 \\
         MultiRC & 3 & pass.: 253.49, q.: 11.70, ans.: 5.84 \\
         \bottomrule
    \end{tabular}
    }%
    \caption{Summary of the evaluated tasks: number of segments ($n$) and average token length of each segment. Measured on training sets.}
    \label{tab:tasks_stats}
    \vspace*{-\baselineskip}
\end{table}

\newpar{AE} \citep{bulian2022answereq}: Answer Equivalence requires determining whether a ``candidate" answer is semantically equivalent to a ``reference" answer, given a question. We use the question and each of the answers as independent text segments, make a binary prediction ``true'' or ``false'', and measure accuracy.

\newpar{BoolQ} \citep{clark2019boolq}: Boolean Questions is a binary question answering task with passages and questions. We use the provided text passage and the question as text segments, and make a binary prediction ``true'' or ``false'', and measure accuracy.

\newpar{BoolQ-Split} A modification of BoolQ, where each passage is split into 5 sub-passages, treated as independent input segments. The sub-passages are formed by greedily merging the passage's sentences, smallest merge first.

\newpar{WiC} \citep{pilehvar2019wic}: Words in Context is a task for evaluating contextual word meanings. Given a word and two sentences in which it occurs, determine whether the word has the same meaning in each sentence. For LAIT, we prefix each sentence by the specified word and treat the newly-prefixed sentences as our text segments. We then predict ``true'' or ``false'', corresponding to whether the word has the same in-context meaning in both sentences. Evaluation is by accuracy.

\newpar{FEVER} \citep{thorne2018fever}: A dataset for fact verification with claims and corresponding evidence. Each claim-evidence pair is labeled as ``supported," ``refuted," or ``NotEnoughInfo." For LAIT, we treat the claim and the evidence as our separate text segments, and aim to predict the correct label. Evaluation is done by accuracy.

\newpar{VitaminC} \citep{schuster2021vitaminc}: A challenging dataset for fact verification which includes ``contrastive evidence", i.e., claim-evidence pairs that differ only slightly (in either the text of the claim or that of the evidence) from another claim-evidence pair, but have a different label. We treat the claim and evidence as independent text segments, and evaluate by accuracy.

\newpar{MultiRC} \citep{MultiRC2018}: The Multi-Sentence Reading Comprehension dataset is a question answering dataset, where each example contains a passage, a question, and an answer\footnote{The original examples have a list of possible answers to each question, but they are split into one example per answer.}. For LAIT, we use the passage, the question, and the answer as the segments. The label is either ``True'' or ``False'' meaning whether the answer is correct or not. Evaluation is done by computing the F1 score over all answers.

\subsection{Baselines}\label{sec:baselines}

We compare LAIT against two groups of baselines: Dual Encoder models and Fully self-attentive models. For the Dual Encoder, we use the SentenceT5-base~\citep{ni-etal-2022-sentence} shared-parameter Dual Encoder which outputs the concatenation of the average of the per-token output representations from the two encoders, together with their difference and dot product, followed by a classifier. We experiment with two depths of classifier: One with a single non-linear layer, and one with 2 additional hidden layers ($d=768$ for all layers). As fully self-attentive baselines, we consider \texttt{T5-base} and \texttt{T5-small} \citep{raffel2020exploring}.

\section{Results}
\label{sec:results}

\begin{figure}[t]
  \includegraphics[width=\linewidth]{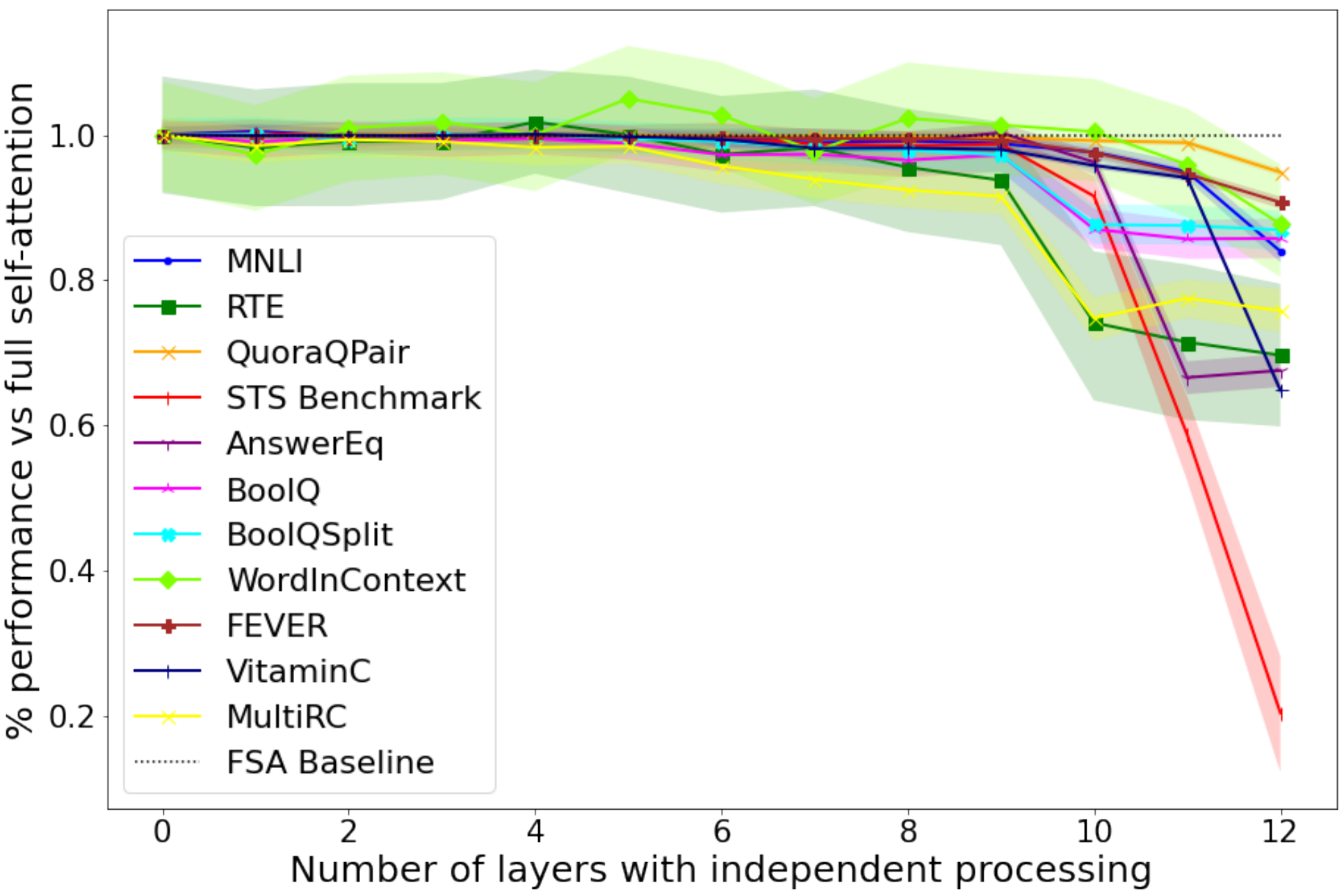}
  \caption{Relative performance of LAIT vs. T5 Fully self-attentive baseline on a variety of multi-segment natural language processing tasks. For all tasks, we report performance on the validation set. Performance only degrades after 8-10 layers of independent segment processing. 95\% confidence interval via bootstrapping on the evaluation data.}
  \label{fig:tasks_scaled}
  \vspace{-10pt}
\end{figure}

\begin{table*}[t]
\resizebox{\linewidth}{!}{%
\begin{tabular}{lcccccccccc}
\toprule
 & MNLI & RTE & QQP & STSB & WiC & BoolQ  & FEVER & VitaminC \\
Model & Accuracy & Accuracy & Accuracy & Spearman  & Accuracy & Accuracy &  Accuracy & Accuracy \\
\midrule
DE + $1\times $ MLP & 75.40 & 51.26 & 90.06 & 24.88 & 61.75 & 69.39 & 86.16 & 56.03   \\
DE + $3\times $ MLP & 77.22 & 56.32 & 89.69 & 62.16 & 60.66 & 69.36 & 87.09 & 65.03  \\
\midrule
T5-small (60M) & 83.42 & 72.92 & 91.14 & 88.67 & 65.83 & 77.92 & 96.57 & 85.35\\
T5-base  (220M) & 86.98 & 84.84 & 91.94 & 90.43 & 72.41 & 83.12 & 97.54 & 88.38\\
\midrule
LAIT-0  & 87.14                         & 80.87             & 91.80             & 90.31                                   & 70.53                       & 82.45                                   & 97.33                       & 88.07 \\
LAIT-1  & 87.14                         & 79.78             & 91.94             & 90.36                                   & 68.65                       & 82.54                                   & 97.25                       & 87.88 \\
LAIT-2  & 86.81                         & 81.59             & 91.87             & 90.19                                   & 70.22                       & 82.39                                   & 97.25                       & 87.89 \\
LAIT-3  & 86.81                         & 79.78             & 91.96             & 89.94                                   & 69.44                       & 82.35                                   & 97.31                       & 87.96 \\
LAIT-4  & 86.84                         & 81.59             & 91.84             & 90.38                                   & 69.59                       & \colorbox{lightgray}{\textbf{82.32}}    & 97.26                       & \colorbox{lightgray}{87.95} \\
LAIT-5  & \colorbox{lightgray}{86.80}   & 79.78             & 91.85             & 89.91                                   & 70.38                       & 80.86                                   & 97.17                       & 87.77 \\
LAIT-6  & 86.23                         & \textbf{80.14}    & 91.79             & 89.63                                   & 71.16                       & 80.86                                   & 97.10                       & \textbf{87.46} \\
LAIT-7  & \textbf{86.29}                & 78.70             & 91.79             & 89.72                                   & 69.44                       & 80.43                                   & 97.07                       & 86.31 \\
LAIT-8  & 86.08                         & 77.98             & 91.55             & \textbf{89.47 }                         & 71.79                       & 80.37                                   & \colorbox{lightgray}{97.05} & 86.49 \\
LAIT-9  & 85.70                         & \underline{\colorbox{lightgray}{78.34}} & \colorbox{lightgray}{91.55}             & \colorbox{lightgray}{\underline{89.39}} & \underline{\textbf{70.85}}  & \underline{80.40}                       & \textbf{96.82}              & 86.26 \\
LAIT-10 & 84.42                         & 61.01             & \textbf{91.07}    & 82.26                                   & \colorbox{lightgray}{67.40} & 71.62                                   & \underline{95.35}           & \underline{84.27} \\
LAIT-11 & \underline{83.00}             & 59.57             & \underline{90.87} & 53.39                                   & 65.05                       & 72.11                                   & 92.13                       & 82.75 \\
LAIT-12 & 73.21                         & 60.29             & 86.85             & 22.68                                   & 59.56                       & 71.50                                   & 88.35                       & 57.00 \\
\bottomrule
\end{tabular}
}%
\caption{Results comparing LAIT configurations with Dual Encoder and Transformer baselines across a variety of sentence-level reasoning tasks. To make comparison easier with other works, we report the best score on the validation set. See Table \ref{tab:synth_dev_table} for a synthetic test-set comparison of LAIT configurations. Most efficient LAIT model within a 99\% performance of LAIT-0 in \textbf{bold}, most efficient LAIT model within 95\% performance of LAIT-0 is \underline{underlined}, most efficient LAIT model where the validation score is within the bootstrapped 95\% confidence interval of LAIT-0 is \colorbox{lightgray}{boxed}.}
\label{tab:dev_table}
\end{table*}

\begin{table}[t]
\resizebox{\linewidth}{!}{%
\begin{tabular}{lccc}
\toprule
 & AnswerEq & BoolQ-split & MultiRC \\
Model & Accuracy & Accuracy & F1 \\
\midrule
T5-small & 89.65 & 77.92 & 73.25 \\
T5-base  & 91.09 & 83.12 & 80.07 \\
\midrule
LAIT-0  & 91.25                       & 81.71 & 78.12                        \\
LAIT-1  & 91.36                       & 82.35 & 77.86 \\
LAIT-2  & 90.46                       & 81.93 & 77.82 \\
LAIT-3  & 90.89                       & 81.53 & \colorbox{lightgray}{77.69} \\
LAIT-4  & 90.85                       & \textbf{82.11} & 77.18 \\
LAIT-5  & 90.78                       & 80.43 & \textbf{77.41} \\
LAIT-6  & 90.62                       & \colorbox{lightgray}{80.76} & \underline{75.60} \\
LAIT-7  & 90.60                       & 79.94 & 73.56 \\
LAIT-8  & 90.06                       & 79.82 & 71.88 \\
LAIT-9  & \colorbox{lightgray}{\textbf{90.98}} & \underline{79.85} & 71.43 \\
LAIT-10 & \underline{87.00}           & 72.20 & 59.50 \\
LAIT-11 & 61.16                       & 71.13 & 61.07 \\
LAIT-12 & 61.02                       & 71.41 & 59.60 \\
\bottomrule
\end{tabular}
}%
\caption{Results for tasks with more than two segments. \textbf{Bold}, \underline{underline}, and \colorbox{lightgray}{box} indicate model performance as in Table \ref{tab:dev_table}.}
\label{tab:dev_table_many_segments}
\vspace{-20pt}
\end{table}

To study the performance-efficiency tradeoff, we consider multiple configurations of LAIT to fully interpolate between a Dual Encoder and a fully self-attentive Transformer. As \texttt{T5-base} has a 12-layer encoder, we consider all LAIT-$p$, for $p \in [0,12]$, where $p$ is the number of layers of independent segment processing before the fully self-attentive component. Note that LAIT-0 is roughly equivalent to \texttt{T5-base}, though it uses a 1-layer decoder vs. the 12-layer decoder of \texttt{T5-base}. 

As can be seen in Tables \ref{tab:dev_table} and \ref{tab:dev_table_many_segments}, which compare best validation-set performance across models, LAIT either matches, nearly-matches, or outperforms the \texttt{T5-base} baseline for every task. This holds even in configurations where cross-segment interaction is delayed to the last few layers of the encoder. As long as there are a few cross-segment interactions later in the model, performance remains relatively stable even as the architecture becomes increasingly efficient; crucially, \textbf{LAIT can delay cross-segment interaction by 8-10 layers without a notable decrease in performance.} We specifically focus on the most efficient LAIT models that: (1) achieve within 99\% of LAIT-0 performance, which we call LAIT-99\%; (2) achieve within 95\% of LAIT-0 perfomrance, called LAIT-95\%; and (3) achieve within the 95\% confidence interval within LAIT-0 performance, called LAIT$\star$. To select these models with higher validity, we perform a synthetic dev/test split of the validation sets and report the held-out validation performance of the LAIT models with the highest performance on the synthetic dev set, reported in Appendix \ref{sec:appendix_additional}.

These results also suggest differences in the proportion of cross-segment processing necessary for different tasks. Sentence and word representation tasks (i.e., Answer Equivalence, STSB, and WiC) have much better LAIT$\star$ models than reasoning-intensive tasks, such as MNLI, BoolQ, and VitaminC. We note that FEVER appears to be easier for LAIT than other ``reasoning" tasks, which we explore further in Section~\ref{sec:robustness}. We also note that some degree of cross-segment processing is necessary for all tasks, evidenced by the steep drop in performance as $p$ approaches $12$ (see Figure~\ref{fig:tasks_scaled}).

\subsection{Scalability}

\begin{table}
    \centering
    \resizebox{\linewidth}{!}{%
    \begin{tabular}{lcc}
        \toprule
        Task & Full Encoding $\downarrow$ & with Caching $\downarrow$ \\
        \midrule
        MNLI     & 66.66\% & 39.71\% \\
        STSB     & 63.07\% & 62.78\% \\
        AnswerEq & 49.94\% & 29.73\% \\
        BoolQ    & 89.72\% & 83.53\% \\
        BoolQ-S  & 42.28\% & 40.51\% \\
        WiC      & 63.45\% & 63.45\% \\
        FEVER    & 72.74\% & 34.68\% \\
        VitaminC & 67.37\% & 41.34\% \\
        MultiRC  & 93.91\% & 50.83\% \\
        RTE      & 92.22\% & 92.02\% \\
        QQP      & 56.06\% & 53.37\% \\
                \midrule
        \multicolumn{3}{l}{\underline{Potential practical settings}:}\\
        ContractNLI & 98.92\% & 21.50\% \\
        WikiClusters & 63.02\% & 16.94\% \\
        \bottomrule
    \end{tabular}
    }%
    \caption{Percent of encoder attention FLOPs (compared to \texttt{T5-base}) when using LAIT-95\% model for each task to process the entire validation set (lower is better). LAIT-95\% selection is based on results in tables \ref{tab:synth_dev_table} and \ref{tab:synth_dev_table_many_segments} in the Appendix.}
    \label{tab:95_flops_table}
    \vspace{-10pt}
\end{table}

\begin{table*}
    \centering
    \begin{tabular}{lllll}
    \toprule
    Dataset & FSA & Sparse & LAIT-12 & LAIT-95\% \\
    \midrule
    MNLI - Full & 167.9 (1.3) & 275.3 (0.96) & 111.3 (1.4) &  116.02 + $\epsilon$ \\
    BoolQ-S - Full & 54.40 (0.43) & 87.72 (0.38) & 37.51 (0.21) & 41.73 + $\epsilon$ \\
    ContractNLI - Single & 0.0071 (0.0012) & 0.0094 (0.0005) & 0.0004 (0.0000) & - \\
    ContractNLI - Full & 25.03 (0.58) & 34.28 (0.46) & 0.0593 (0.0008) & - \\
    WikiClusters - Single & 1390. (6.0) & 1871. (7.1) & 1.086 (0.03) & - \\
    WikiClusters - Full & 4805. (32.) & 5451. (15.) & 87.79 (0.78) & - \\
    \bottomrule
    \end{tabular}
    \caption{Encoding latency (and standard deviation) comparison in seconds between fully self-attentive T5 (FSA), LongT5 with local attention (Sparse), and LAIT. $\epsilon$ represents system-dependent processing of a LAIT cache. Measurements were performed with a 2080Ti GPU, using the Hugging Face \cite{wolf2019huggingface} implementation of T5 and LongT5. }
    \label{tab:latency}
    \vspace{-10pt}
\end{table*}

By deferring the expensive cross-segment attention to later stages of the model, LAIT both reduces the attention complexity of the model, and enables the caching and reuse of partial representations computed before the cross-segment attention layers.

Table \ref{tab:95_flops_table} shows improvements in attention FLOPs for LAIT, both with and without caching of the intermediate representations, when using the LAIT-95\% model. Table \ref{tab:star_flops_table} contains results for LAIT$\star$. As we would expect from Equation \ref{eq:op_formula}, datasets with text segments of similar size benefit the most in the typical setting. Howevever, to fully realize this benefit for single forward passes would require a custom kernel, such as those implemented in work on sparse transformers.

\subsection{Caching and Reusing Representations}

A key advantage of the delayed cross-segment interaction in LAIT is the ability to cache and reuse intermediate representations of text segments. Unlike in benchmarks, real-world settings almost never process a set of segments in isolation; it is much more likely that the processing of a set of text segments occurs as part of a larger task such as document comparison, document analysis, or claim verification.

Recently, a number of datasets \citep{schuster2022stretching, koreeda2021contractnli, petroni2022wikipedia} have suggested the usefulness of natural language inference in large-scale real-world reasoning tasks. In one such dataset, ContractNLI \citep{koreeda2021contractnli}, a fixed set of 17 claims are evaluated against different legal contracts. In other scenarios \citep{schuster2022stretching, gu2020generating}, the contents of multiple documents within a cluster of related documents must be compared.

In both scenarios, a typical approach would require comparing each sentence within a document with each other sentence, leading to a complexity that scales quadratically with the size of the document cluster, the size of the documents, and the length of the sentences. But with LAIT, the bulk of the work will be performed only once. Because each document or claim can be encoded independently for most of the layers of the model, the latency improvement offered by LAIT in these settings is related to the overall redundancy and duplication of text segments within the task.

Table \ref{tab:latency} demonstrates the savings possible for both popular academic tasks, and two realistic settings: ContractNLI~\citep{koreeda2021contractnli}, and WikiClusters \citep{schuster2022stretching}. For MNLI and BoolQ, we measure the time to encode the entire dataset. For WikiClusters and ContractNLI, we both measure the time to encode the entire dataset and the time to encode a single document (in the case of ContractNLI) or cluster (in the case of WikiClusters). We compare a standard fully self-attentive model (T5), a sparse model (LongT5 with local attention), and LAIT. For MNLI and BoolQ, we estimate the latency of the LAIT-95\% model for that task, as a weighted average of FSA and LAIT layers.

Even without a custom kernel, LAIT's independent processing of input segments enables significant speedups for processing real-world data. Interestingly, the sparse transformer demonstrates slightly \textit{increased} latency, likely because the the input sizes are relatively short. However, even when enabled by a sparse transformer, processing larger chunks of data -- such as an entire ContractNLI contract alongside each of the 17 claims -- will not fully alleviate the problem, as the contracts must still be processed 17 times, rather than just once as in LAIT. In these situations, LAIT may be able to complement a sparse transformer; this would require further study.

\subsection{Robustness}
\label{sec:robustness}

\begin{table}[t]
    \centering
    \begin{tabular}{lcccc}
        \toprule
         Model & FEVER & VitaminC &  MNLI \\
        \midrule
         \multicolumn{4}{l}{\underline{Training Data: FEVER-train}} \\
         LAIT-0  & 97.33 & 65.12 & 47.93 \\
         LAIT-3  & 97.31 & 64.73 & 45.85 \\
         LAIT-6  & 97.10 & 63.62 & 35.15 \\
         LAIT-9  & 96.82 & 62.97 & 33.82 \\
         LAIT-12 & 88.35 & 49.91 & 34.29 \\ %
                  \midrule
         \multicolumn{4}{l}{\underline{Training Data: VitaminC-train}} \\
         LAIT-0 & 78.54 & 88.07 & 80.37 \\
         LAIT-3 & 78.96 & 87.96 & 80.01 \\
         LAIT-6 & 78.72 & 87.46 & 78.74 \\
         LAIT-9 & 77.70 & 86.26 & 76.74 \\
         LAIT-12 & 54.04 & 57.00 & 43.38 \\
         \bottomrule
    \end{tabular}
    \caption{Accuracy of FEVER- and VitaminC-trained LAIT models on FEVER, VitaminC, and MNLI.}
    \vspace{-10pt}
    \label{tab:robust}
\end{table}

A potential concern with an approach like LAIT is that it may be more susceptible to reported biases in sentence-level models \citep{poliak-etal-2018-hypothesis, schuster2021vitaminc}. We test LAIT's effect on the model's robustness to domain shifts, and to biases in the training data such as over-relying on clues in one of the segments instead of performing cross-segment reasoning.

\citet{schuster2021vitaminc} found that in FEVER, when evidence text in a claim-evidence pair was revised in a way that would reverse the semantic relationship (e.g., $f_{\textnormal{revision}}$(Claim, Evidence, \textsc{Refutes}) $\rightarrow$ (Claim, Evidence$'$, \textsc{Supports}), models trained on FEVER would only make the correct prediction 56\% of the time. Table \ref{tab:robust} summarizes our robustness experiments using zero-shot transfer from FEVER and VitaminC.

We find that when LAIT is trained on on FEVER, the transfer performance drops faster than the in-domain performance as independence is increased. However, when training on VitaminC, the decrease in accuracy as a function of $P$ is more correlated with the in-domain trend. This suggests that LAIT models can be robust against domain shifts and contrastive adversarial examples when trained with appropriate data.

\section{Related Work}
\label{sec:related}

\newpar{Sentence encoders.}
Modern representation learning systems at the sentence level have rapidly risen in popularity, starting with InferSent \citep{conneau-etal-2017-supervised}, ESIM \citep{cer-etal-2018-universal}, and USE \citep{chen-etal-2017-enhanced}. Following the inception of Transformer~\citep{transformer}, new sentence encoders~\cite[see e.g.,][]{gao-etal-2021-simcse,ni-etal-2022-sentence,reimers-2019-sentence-bert} demonstrated improved performance on many sentence-pair benchmarks. Other work extended this approach to document encoders by hierarchically encoding sentences independently before combining them into a pooled document embedding~\citep{wu-etal-2021-hi,Yang_2020}. Yet, unlike previous work, LAIT effectively breaks a pretrained Transformer into a hybrid of multiple parallel segment encoders and powerful fully-attentive layers to match state-of-the-art performance across many NLP tasks.

\newpar{Efficient text classifiers}
Dual encoder architectures, originally dating back to the Siamese architecture of \citep{bromley94}, were proposed for efficient retrieval in \citep{gillick2018end}. \citep{ni-2021-gtr-arxiv} and  \citep{pmlr-v162-menon22a} significantly broaden the range of tasks efficiently served by dual encoders.

Building on the Transformer architecture, LAIT can also readily leverage many other known efficiency solutions~\citep{tay2022survey} such as distillation \citep{sanh2019distilbert, jiao2020tinybert}, quantization \citep{shen2020qbert, zafrir2019q8bert}, and early exiting \citep{schuster2022calm, xin2020deebert}. 

\newpar{Sparse attention.} Sparse attention architectures have demonstrated that not all attention connections within a Transformer are necessary, and that impressive performance can be achieved even when removing a large number of the cross-token attention. Examples such as BigBird, Longformer, and LongT5 \citep{zaheer2020bigbird, beltagy2020longformer, guo2021longt5} use local attention windows and some form of global attention to reduce the attention complexity. Other approaches dynamically skip certain computations \citep{sinkhorn}. Unlike these approaches, here we impose the sparsity on top of known input segments, which preserves segment-level semantics and supports parallel computing and caching of segments. Despite their benefits, sparse transformers still include cross-segment attention at every layer of the model, and as such they cannot encode segments independently.

\newpar{Late interaction.} 
Some recent work has considered precomputing full-token representations of some, but not all, text segments, as well as late interaction between queries and documents~\citep{lu2020twinbert,xiong2017dynamic}. ColBERT \citep{khattab2020colbert, santhanam-etal-2022-colbertv2} uses precomputed token representations as part of a DE retrieval framework. %
These architectures, however, are tailored for retrieval tasks that use embedding similarity scores, and generally under-perform in classification tasks like NLI. The fully-attentive layers in LAIT allow bridging this performance gap while still providing efficiency gains.
Our caching variant also relates to other recent parallel work on precomputing and reusing representations of repeated passages to speed up computation \citep{saadfalcon2023embedding,dejong2023precomputed,lidecoupled}.
\citet{hui-etal-2022-ed2lm} develop a fully parallel encoder for documents and queries, where both encodings are fed to a joint decoder for re-ranking.
Most similar to our work is \citet{MacAvaney_2020} that study a hybrid Transformer architecture for ranking. In this work, we focus on general NLP tasks with an arbitrary number of segments, and unconstrained output space.

\section{Conclusion}
\label{sec:conclusion}
We present Layer-Adjustable Interactions in Transformers (LAIT) to allow simple-but-effective efficiency gains over a wide range of NLP tasks. The LAIT framework leverages existing pretrained Transformers such as T5, and converts them during finetuning into a hybrid model that combines parallel independent encoding of multiple segments, followed by fully-attentive layers to allow cross-segment reasoning.

We evaluate LAIT on a large set of 10 well-known datasets, involving different examined capabilities, number of segments, input lengths, output spaces, and difficulty levels. We find LAIT to consistently provide significant reduction in encoder attention complexity while preserving high accuracy. Furthermore, we show that the parallel independent segment encoding of LAIT enables additional inference-time compute savings by caching representations of repeated segments in large scale real-world settings. 

LAIT demonstrates that transformers can achieve high performance even without cross-segment interaction at every layer; essentially, that sentences can be just as effectively encoded if first processed separately, and then processed jointly.

\section*{Limitations}
\label{sec:limitations}

While the LAIT framework can significantly reduce the computation required for large-scale sentence-level reasoning and classification tasks, we do foresee some limitations in its use. Caching per-token representations for large numbers of text segments leads to a dramatic increase in memory requirements, which could be prohibitive for extremely low-compute end users. We also note that LAIT can further exacerbate segment-level bias in datasets. While we believe that careful data curation approaches ameliorate this issue, the risk of bias is not always known to downstream users and as such corrective datasets may not always be available. Finally, LAIT can increase the cost of training because the optimal degree of independence is not known until all LAIT-$p$ models are evaluated, though in practical settings (1) it is possible to perform a binary search of LAIT configurations because performance generally decreases monotonically as $p$ increases; (2) even a naive rule of setting $p$ to a quarter of the model's depth seems to provide some immediate gains while preserving 99\% of the accuracy in all our evaluated tasks; and (3) inference-time cost improvements will far outweigh training costs.

\bibliography{anthology,custom}

\appendix
\newpage
\section{Segment Preprocessing}
\label{sec:concat_strategy}

For each task, we must prepare the text segments for processing by either the Dual Encoder, Fully Self-attentive, or LAIT models. Here, we report the preprocessing and concatenation strategy used. For the FSA models, we concatenate each segment. For the DE and LAIT models, we treat each segment as a separate input.

\paragraph{MNLI} 
\begin{itemize}
    \item \texttt{hypothesis: <hypothesis text>}
    \item \texttt{premise: <premise text>}
\end{itemize}

\paragraph{WiC}
\begin{itemize}
    \item \texttt{<key word>: <sentence1>}
    \item \texttt{<key word>: <sentence2>}
\end{itemize}
 
\paragraph{STSB}
\begin{itemize}
    \item \texttt{sentence1: <sentence1>}
    \item \texttt{sentence2: <sentence2>}
\end{itemize}

\paragraph{BoolQ}
\begin{itemize}
    \item \texttt{question: <question>}
    \item \texttt{passage: <passage>}
\end{itemize} 

\paragraph{RTE}
\begin{itemize}
    \item \texttt{hypothesis: <hypothesis>}
    \item \texttt{premise: <premise>}
\end{itemize} 

\paragraph{QQP}
\begin{itemize}
    \item \texttt{question1: <question1>}
    \item \texttt{question2: <question2>}
\end{itemize} 

\paragraph{FEVER}
\begin{itemize}
    \item \texttt{hypothesis: <claim>}
    \item \texttt{premise: <evidence>}
\end{itemize} 

\paragraph{VitaminC}
\begin{itemize}
    \item \texttt{hypothesis: <claim>} 
    \item \texttt{premise: <evidence>} 
\end{itemize}

\paragraph{Answer Equivalence}
\begin{itemize}
    \item \texttt{question: <question>} 
    \item \texttt{answer1: <answer1>}
    \item \texttt{answer2: <answer2>}
\end{itemize}

\paragraph{MultiRC}
\begin{itemize}
    \item \texttt{question: <question>}
    \item \texttt{answer: <answer>}
    \item \texttt{paragraph: <paragraph>}
\end{itemize}

\paragraph{BoolQ-Split}
\begin{itemize}
    \item \texttt{question: <question>} 
    \item \texttt{passage1: <passage1>}
    \item \texttt{passage2: <passage2>}
    \item \texttt{passage3: <passage3>}
    \item \texttt{passage4: <passage4>}
    \item \texttt{passage5: <passage5>}
\end{itemize}

\newpage

\section{Additional Results}
\label{sec:appendix_additional}

\subsection{Full Decoder and T5-Large Models}
\label{sec:appendix_larger_models}
For our experiments in the main paper we used a T5-base model with only a single decoder layer. Using only one decoder layer is faster at inference time enforces the model to more heavily rely on the encoder stack, and therefore the strong results of LAIT in that setting are even more encouraging. We also experiment with a LAIT on top of a T5-Base with all 12 decoder layers and with a larger T5-Large that has 24 layers in both encoder and decoder stacks. 

Table~\ref{tab:full_dec} and Table~\ref{tab:large} present the results for T5-Base and T5-Large, respectively. LAIT shows similar trends for these different configurations, indicating that our approach is general and translates to different model configurations. Also, as expected, larger decoder allows LAIT to further postpone the cross-segment interactions (larger $P$) without loosing accuracy.

\begin{table*}[t]
    \centering
    \resizebox{\linewidth}{!}{%
    \begin{tabular}{lccccccccccc}
    \toprule
        $P$ & \multicolumn{2}{c}{MNLI} & \multicolumn{2}{c}{QQP} & \multicolumn{2}{c}{WiC} & \multicolumn{2}{c}{WiC} & \multicolumn{2}{c}{MultiRC}\\ 
        ~ &  Accuracy & Relative &  Accuracy & Relative &  Accuracy & Relative &  Accuracy & Relative &  F1 & Relative \\
        \midrule
0        & 86.92        &                   & 91.86        &          & 72.73        &           & 83.64        &           & 80.26        &           \\
1        & 86.90         & 99.98\%           & 91.89        & 100.03\%          & 72.57        & 99.78\%           & 83.46        & 99.78\%           & 79.74        & 99.35\%           \\
2        & 87.05        & 100.15\%          & 91.90         & 100.04\%          & 72.88        & 100.21\%          & 83.49        & 99.82\%           & 79.95        & 99.61\%           \\
3        & 87.17        & 100.29\%          & 91.93        & 100.08\%          & 73.51        & 101.07\%          & 83.49        & 99.82\%           & 79.80         & 99.43\%           \\
4        & 86.93        & 100.01\%          & 91.87        & 100.01\%          & 72.88        & 100.21\%          & 83.64        & 100.00\%          & 79.69        & 99.29\%           \\
5        & 86.60         & 99.63\%           & 91.94        & 100.09\%          & 73.51        & 101.07\%          & 83.15        & 99.41\%           & 78.92        & 98.33\%           \\
6        & 86.61        & 99.64\%           & 91.72        & 99.85\%           & 73.67        & 101.29\%          & 82.97        & 99.20\%           & 78.37        & 97.65\%           \\
7        & 86.30         & 99.29\%           & 91.66        & 99.78\%           & 73.82        & 101.50\%          & 82.45        & 98.58\%           & 78.03        & 97.22\%           \\
8        & 86.15        & 99.11\%           & 91.73        & 99.86\%           & 73.67        & 101.29\%          & 82.48        & 98.61\%           & 78.13        & 97.35\%           \\
9        & 86.13        & 99.09\%           & 91.61        & 99.73\%           & 73.82        & 101.50\%          & 82.35        & 98.46\%           & 77.96        & 97.13\%           \\
10       & 84.97        & 97.76\%           & 91.45        & 99.55\%           & 71.32        & 98.06\%           & 77.13        & 92.22\%           & 67.07        & 83.57\%           \\
11       & 84.17        & 96.84\%           & 90.98        & 99.04\%           & 67.87        & 93.32\%           & 74.74        & 89.36\%           & 59.06        & 73.59\%           \\
12       & 83.22        & 95.74\%           & 89.55        & 97.49\%           & 64.89        & 89.22\%           & 73.73        & 88.15\%           & 58.18        & 72.49\%       \\   
        \bottomrule
    \end{tabular}
    }%
    \caption{Results for different number of parallel layers $P$ of LAIT using the same setting as Table~\ref{tab:dev_table}, but with 12 decoder layers instead of a single decoder layer. Hence, $P=0$ is similar to T5-base setting from Table~\ref{tab:dev_table} (numbers are not identical due to different training runs). The extra decoding layers allows further increasing $P$ compared to single decoder-layer while maintaining similar performance. The relative column shows the accuracy or F1 change compared to $P=0$.}
\label{tab:full_dec}
\end{table*}

\begin{table*}[t]
    \centering
    \begin{tabular}{lcccccccc}
    \toprule
        $P$ & \multicolumn{2}{c}{MNLI} & \multicolumn{2}{c}{WiC} & \multicolumn{2}{c}{BoolQ} & \multicolumn{2}{c}{MultiRC}\\ 
        ~ &  Accuracy & Relative &  Accuracy & Relative &  Accuracy & Relative &  F1 & Relative\\
        \midrule
0  & 90.19 &          & 73.82 &          & 86.88 &          & 84.16   &          \\
1  & 90.01 & 99.80\%  & 73.35 & 99.36\%  & 86.88 & 100.00\% & 84.03   & 99.85\%  \\
2  & 90.16 & 99.97\%  & 73.82 & 100.00\% & 86.76 & 99.86\%  & 83.76   & 99.52\%  \\
3  & 90.10  & 99.90\%  & 73.35 & 99.36\%  & 86.85 & 99.97\%  & 84.04   & 99.86\%  \\
4  & 89.97 & 99.76\%  & 73.51 & 99.58\%  & 87.25 & 100.43\% & 84.20    & 100.05\% \\
5  & 90.09 & 99.89\%  & 74.14 & 100.43\% & 87.19 & 100.36\% & 84.26   & 100.12\% \\
6  & 89.97 & 99.76\%  & 74.29 & 100.64\% & 87.09 & 100.24\% & 84.19   & 100.04\% \\
7  & 90.39 & 100.22\% & 74.14 & 100.43\% & 87.22 & 100.39\% & 83.75   & 99.51\%  \\
8  & 90.15 & 99.96\%  & 74.45 & 100.85\% & 86.88 & 100.00\% & 84.04   & 99.86\%  \\
9  & 90.07 & 99.87\%  & 73.98 & 100.22\% & 87.22 & 100.39\% & 83.86   & 99.64\%  \\
10 & 89.87 & 99.65\%  & 74.29 & 100.64\% & 86.94 & 100.07\% & 84.00      & 99.81\%  \\
11 & 89.84 & 99.61\%  & 74.45 & 100.85\% & 87.03 & 100.17\% & 83.82   & 99.60\%  \\
12 & 90.13 & 99.93\%  & 74.92 & 101.49\% & 87.06 & 100.21\% & 83.97   & 99.77\%  \\
13 & 89.75 & 99.51\%  & 74.29 & 100.64\% & 86.88 & 100.00\% & 83.54   & 99.26\%  \\
14 & 89.59 & 99.33\%  & 73.82 & 100.00\% & 86.45 & 99.51\%  & 83.11   & 98.75\%  \\
15 & 89.86 & 99.63\%  & 72.73 & 98.52\%  & 86.94 & 100.07\% & 82.80    & 98.38\%  \\
16 & 89.81 & 99.58\%  & 73.04 & 98.94\%  & 86.70  & 99.79\%  & 82.44   & 97.96\%  \\
17 & 89.50  & 99.23\%  & 73.98 & 100.22\% & 86.09 & 99.09\%  & 81.85   & 97.26\%  \\
18 & 89.37 & 99.09\%  & 73.51 & 99.58\%  & 86.02 & 99.01\%  & 81.57   & 96.92\%  \\
19 & 88.66 & 98.30\%  & 74.14 & 100.43\% & 84.89 & 97.71\%  & 78.99   & 93.86\%  \\
20 & 88.50  & 98.13\%  & 72.88 & 98.73\%  & 83.33 & 95.91\%  & 76.66   & 91.09\%  \\
21 & 88.39 & 98.00\%  & 73.82 & 100.00\% & 82.45 & 94.90\%  & 74.67   & 88.72\%  \\
22 & 88.16 & 97.75\%  & 72.26 & 97.89\%  & 81.83 & 94.19\%  & 73.02   & 86.76\%  \\
23 & 86.93 & 96.39\%  & 71.16 & 96.40\%  & 79.24 & 91.21\%  & 61.11   & 72.61\%  \\
24 & 85.83 & 95.17\%  & 68.03 & 92.16\%  & 76.88 & 88.49\%  & 59.34   & 70.51\% \\
        \bottomrule
    \end{tabular}
    \caption{Results for different number of parallel layers $P$ of LAIT with a T5-Large backbone model, using all 24 decoder layers.  The relative column shows the accuracy or F1 change compared to $P=0$.}
\label{tab:large}
\end{table*}

\subsection{Generalization of LAIT configuration}
\label{sec:appendix_test_set}

\begin{table*}
\resizebox{\linewidth}{!}{%
\begin{tabular}{lcccccccc}
\toprule
 & MNLI & RTE & QQP & STSB & WiC & BoolQ  & FEVER & VitaminC \\
Model & Accuracy & Accuracy & Accuracy & Spearman & Accuracy & Accuracy &  Accuracy & Accuracy \\
\midrule
LAIT-0  & 86.86$\pm$ 0.93 & \textbf{78.42 $\pm$ 6.47} & 91.57 $\pm$ 0.37 & 89.75$\pm$ 1.64 & 68.97$\pm$ 5.17 & 81.65$\pm$ 1.87 & 97.01$\pm$ 0.44 & 87.95$\pm$ 0.36 \\
LAIT-1  & 86.86$\pm$ 0.94 & 71.94 $\pm$ 7.19 & 91.61 $\pm$ 0.39 & 89.53$\pm$ 1.68 & 67.08$\pm$ 5.17 & 81.96$\pm$ 1.81 & 96.90$\pm$ 0.46 & 87.80$\pm$ 0.37 \\
LAIT-2  & 86.37$\pm$ 0.94 & 76.26 $\pm$ 6.51 & 91.43 $\pm$ 0.37 & 89.24$\pm$ 1.82 & 68.34$\pm$ 5.02 & 81.59$\pm$ 1.87 & 96.92$\pm$ 0.45 & 87.83$\pm$ 0.37 \\
LAIT-3  & 86.29$\pm$ 0.93 & 74.1 $\pm$ 7.19 & 91.87 $\pm$ 0.35 & 88.91$\pm$ 1.85 & 66.77$\pm$ 5.49 & 81.41$\pm$ 1.93 & 96.94$\pm$ 0.44 & 87.87$\pm$ 0.34 \\
LAIT-4  & 86.43$\pm$ 0.93 & \underline{76.98 $\pm$ 6.47} & 91.64 $\pm$ 0.37 & 89.67$\pm$ 1.63 & 68.03$\pm$ 5.49 & \textbf{81.59$\pm$ 1.81} & 97.01$\pm$ 0.44 & 87.92$\pm$ 0.34 \\
LAIT-5  & \colorbox{lightgray}{\textbf{86.51$\pm$ 0.93}} & 74.1 $\pm$ 7.19 & 91.65 $\pm$ 0.38 & \textbf{88.99$\pm$ 1.88} & 68.65$\pm$ 5.18 & 79.82$\pm$ 1.87 & 96.71$\pm$ 0.45 & \colorbox{lightgray}{87.73$\pm$ 0.36} \\
LAIT-6  & 85.84$\pm$ 1.01 & 70.5 $\pm$ 7.23 & 91.53 $\pm$ 0.4 & 88.73$\pm$ 1.79 & 68.34$\pm$ 5.02 & 80.49$\pm$ 1.93 & 96.68$\pm$ 0.45 & \textbf{87.41$\pm$ 0.34} \\
LAIT-7  & 85.94$\pm$ 0.91 & 74.1 $\pm$ 7.19 & 91.37 $\pm$ 0.4 & 88.82$\pm$ 1.82 & 66.46$\pm$ 5.02 & 80.06$\pm$ 1.87 & 96.68$\pm$ 0.47 & 86.21$\pm$ 0.39 \\
LAIT-8  & 85.80$\pm$ 1.00 & 72.66 $\pm$ 7.19 & 91.4 $\pm$ 0.39 & 88.60$\pm$ 1.85 & \textbf{70.53$\pm$ 4.86} & 79.57$\pm$ 1.99 & \colorbox{lightgray}{96.70$\pm$ 0.47} & 86.35$\pm$ 0.39 \\
LAIT-9  & 85.19$\pm$ 0.99 & \colorbox{lightgray}{72.66 $\pm$ 7.19} & \colorbox{lightgray}{91.44 $\pm$ 0.38} & \colorbox{lightgray}{\underline{88.38$\pm$ 1.79}} & \underline{67.08$\pm$ 5.17} & \colorbox{lightgray}{\underline{80.37$\pm$ 1.96}} & \textbf{96.52$\pm$ 0.48} & 86.18$\pm$ 0.39 \\
LAIT-10 & \underline{83.80$\pm$ 1.01} & 52.52 $\pm$ 7.91 & \textbf{90.89 $\pm$ 0.42} & 79.21$\pm$ 2.90 & 64.26$\pm$ 5.02 & 70.40$\pm$ 2.23 & \underline{94.80$\pm$ 0.54} & \underline{84.00$\pm$ 0.42} \\
LAIT-11 & 82.17$\pm$ 1.10 & 53.24 $\pm$ 7.91 & \underline{90.33 $\pm$ 0.41} & 51.49$\pm$ 5.44 & \colorbox{lightgray}{65.20$\pm$ 4.86} & 70.89$\pm$ 2.05 & 91.48$\pm$ 0.70 & 82.60$\pm$ 0.42 \\
LAIT-12 & 72.19$\pm$ 1.27 & 51.08 $\pm$ 7.91 & 86.81 $\pm$ 0.47 & 18.30$\pm$ 6.94 & 58.93$\pm$ 5.49 & 70.58$\pm$ 2.08 & 88.14$\pm$ 0.83 & 57.00$\pm$ 0.54 \\
\bottomrule
\end{tabular}
}%
\caption{Results comparing LAIT configurations. We perform a split of the validation sets to form synthetic validation and test sets; we report the test-set score corresponding to the checkpoint with the best validation performance.}
\label{tab:synth_dev_table}
\end{table*}

Here, we report additional results using our split of the existing validation sets into a synthetic validation set and a heldout test set.

\begin{figure}[h]
  \includegraphics[width=\linewidth]{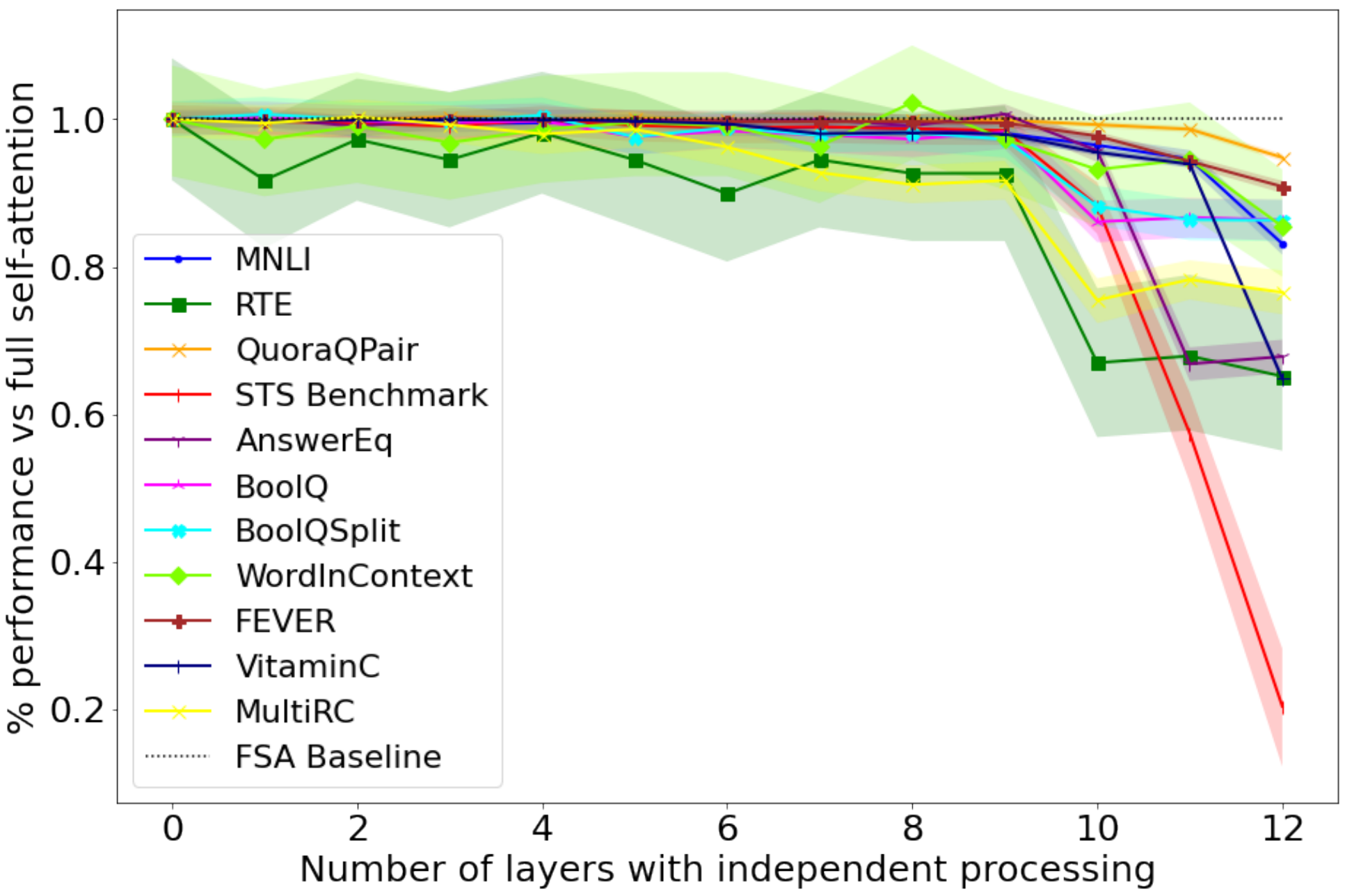}
  \caption{Relative performance of LAIT vs. T5 Fully self-attentive baseline on a variety of multi-segment natural language processing tasks. For all tasks, we report performance on a held-out portion of the validation set. Performance only degrades after 8-10 layers of independent segment processing. 95\% confidence interval via bootstrapping on the evaluation data.}
  \label{fig:tasks_scaled_synth}
\end{figure}

Figure \ref{fig:tasks_scaled_synth} reports the decrease in model performance as the number of parallel encoder layers increases.
Table \ref{tab:synth_dev_table} reports the heldout test results for the LAIT models with the best synthetic validation performance.
Table \ref{tab:synth_dev_table_many_segments} includes the tasks with more than two segments.
Table \ref{tab:star_flops_table} reports the cost of both full encoding and partially-cached encoding for LAIT$\star$ models identified from Tables \ref{tab:synth_dev_table} and \ref{tab:synth_dev_table_many_segments}.

\begin{table}[t]
    \centering
    \begin{tabular}{lcc}
        \toprule
        Task & Full Encoding $\downarrow$ & with Caching $\downarrow$ \\
        \midrule
        MNLI     & 83.33\% & 69.85\% \\
        STSB     & 63.07\% & 62.78\% \\
        AnswerEq & 54.94\% & 41.21\% \\
        BoolQ    & 89.72\% & 83.53\% \\
        BoolQ-S  & 48.69\% & 47.12\% \\
        WiC      & 55.85\% & 55.85\% \\
        FEVER    & 78.19\% & 47.74\% \\
        VitaminC & 80.42\% & 70.67\% \\
        MultiRC  & 94.93\% & 59.02\% \\
        RTE      & 82.49\% & 82.04\% \\
        QQP      & 64.05\% & 61.85\% \\
        \\
        \multicolumn{3}{l}{Potential practical settings:}\\
        \midrule
        ContractNLI & 99.46\% & 60.75\% \\
        WikiClusters & 81.51\% & 58.47\% \\
        \bottomrule
    \end{tabular}
    \caption{Cost of encoder attention FLOPs (vs. \texttt{T5-base}) when using LAIT$\star$ model for each task to process the entire validation set. LAIT$\star$ selection is based on results in tables \ref{tab:synth_dev_table} and \ref{tab:synth_dev_table_many_segments}.}
    \label{tab:star_flops_table}
\end{table}

\begin{table}
\resizebox{\linewidth}{!}{%
\begin{tabular}{lccc}
\toprule
 & AnswerEq & BoolQ-split & MultiRC \\
Model & Accuracy & Accuracy & F1 \\
\midrule
LAIT-0  & 90.55$\pm$ 1.21                       & 81.22$\pm$ 1.90 & 78.55 $\pm$ 1.94                         \\
LAIT-1  & 90.73$\pm$ 1.17                      & 81.77$\pm$ 1.87 & 78.13 $\pm$ 1.91 \\
LAIT-2  & 89.65$\pm$ 1.19                       & 81.16$\pm$ 1.90  & 78.81 $\pm$ 1.89 \\
LAIT-3  & 90.69$\pm$ 1.19                       & 81.04$\pm$ 1.96 & \textbf{77.97} $\pm$ 1.9 \\
LAIT-4  & 90.55$\pm$ 1.22                       & \textbf{81.65$\pm$ 1.83} & 76.98 $\pm$ 2.15 \\
LAIT-5  & 90.46$\pm$ 1.21                       & 79.27$\pm$ 1.90 & \colorbox{lightgray}{77.48 $\pm$ 1.97} \\
LAIT-6  & 90.37$\pm$ 1.19                       & 80.31$\pm$ 1.96  & \underline{75.6} $\pm$ 2.12 \\
LAIT-7  & 90.51$\pm$ 1.22                       & 79.45$\pm$ 1.96 & 72.87 $\pm$ 1.99 \\
LAIT-8  & 89.74$\pm$ 1.26                       &  \colorbox{lightgray}{79.76$\pm$ 1.99} & 71.58 $\pm$ 1.97 \\
LAIT-9  & \colorbox{lightgray}{\textbf{91.18$\pm$ 1.15}} & \underline{79.02$\pm$ 2.02}  & 72.03 $\pm$ 2.04  \\
LAIT-10 & \underline{86.68$\pm$ 1.44}           & 71.62$\pm$ 2.11 & 59.29 $\pm$ 2.46  \\
LAIT-11 & 60.50$\pm$ 1.91                       & 70.15$\pm$ 2.20 & 61.47 $\pm$ 2.13  \\
LAIT-12 & 61.40$\pm$ 2.02                       & 70.09$\pm$ 2.11 & 60.11 $\pm$ 2.34  \\
\bottomrule
\end{tabular}
}%
\caption{Results for tasks with more than two segments. \textbf{Bold}, \underline{underline}, and \colorbox{lightgray}{box} indicate model performance as in Table \ref{tab:synth_dev_table}.}
\label{tab:synth_dev_table_many_segments}
\end{table}

\end{document}